\documentclass{article}

\usepackage[preprint]{neurips_2020}

\usepackage[utf8]{inputenc} 
\usepackage[T1]{fontenc}    
\usepackage{hyperref}       
\usepackage{url}            
\usepackage{booktabs}       
\usepackage{amsfonts}       
\usepackage{nicefrac}       
\usepackage{microtype}      
\usepackage{ulem}
\usepackage{graphicx}

\title{Towards a universal language of concepts: A survey}
\author{Aishni Parab\\
  Department of Computer Science\\
  UCLA\\
  \texttt{aishni@g.ucla.edu} \\
}

\begin{document}

\maketitle

\begin{abstract}
 Humans can learn and generalize novel concepts from sparse data because they express knowledge in rich structural formats. In this paper, we propose that programs are a strong candidate for universal representation of concepts. We review computational models of concept learning that use programs as their concept representation and evaluate their contribution toward a universal representational language.
\end{abstract}

\section*{Author's note}
This work was originally completed as a capstone project for the
M.S. in Computer Science at the University of California, Los Angeles,
in 2022.

\section{Introduction}
Concepts are fundamental building blocks of thought. To arbitrarily imagine a cow with pink zebra stripes, one must first access the concepts of a cow, the color pink, and a zebra, then compose them in a meaningful form. But how does the human mind represent this form? Thoughts by nature are unbounded; the cow with pink zebra stripes can morph into a brown cow with yellow stripes or grow two additional tails. Then what representation will allow these transformations?

Such gradation of thought is also prevalent in more rational ubiquitous contexts such as planning, in which humans simulate a sequence of actions to achieve a goal \cite{bratman1987intention}. For instance, if a student intends to complete their homework assignment today, they may plan to visit the library to collect a relevant reference book. This plan now includes collecting the reference book as a subgoal, which can become further involved if we account for reserving the book online before picking it up as a sub-subgoal. This plan requires an understanding of the concepts of homework, assignment, deadline, book, library, reservation, trip, and item collection. How does the human mind support such reasoning over simple and complex concepts despite having limited cognitive resources like working memory and speed \cite{lieder2020resource}?

To easily manipulate concepts in sophisticated ways, the human mind must represent concepts in a form that can be readily conceived, stored, retrieved, processed, and updated. The following sections evaluate what is required from a mental representation to represent concepts efficiently and then argue that programs are a feasible universal representation of this goal. We define the meaning of a program representation,  highlight its unique properties and evaluate whether a universal representation of concepts is enough to achieve complex reasoning. Finally, we review various computational models of concept learning that use programs as their concept representation and assess their contribution toward a universal language of concepts. 

\section{Needed properties of a concept representation}

To identify an ideal mental representation of concepts, we evaluate what the human mind can do with concepts and determine the properties this demands from a representation. 

Human infants are capable of learning new concepts from sparse and noisy data. Experiments have shown that children as young as three years of age can learn the meaning of a new word with just one example \cite{carey1978acquiring}. They can also build higher-order generalizations about concepts \cite{sim2017learning}. For example, children attribute the concept that "dogs bark" to generalize that all animals of the same kind make the same sound \cite{shipley1993categories}. Studies support that 9-month-old infants can do this using very few examples \cite{dewar2010induction}. 

Infants can make such generalizations because they form a hypothesis that they apply to novel instances, commonly referred to as "overhypothesis" \cite{goodman1983fact}. Several authors have alternately referred to this ability as the formation of "intuitive theories" or "schemata," but we can generally interpret it as "abstract knowledge" \cite{ullman2012theory, tenenbaum2010intuitive, xu2021bayesian, carey2000origin, wellman1992cognitive, dewar2010induction, kemp2008discovery}. 

This abstract knowledge, typically referred to as "inductive bias" in the artificial intelligence community or "priors" in the statistics community, constrains the learners' search space of possible solutions and takes the form of rich a structural representation 
\cite{tenenbaum2011grow, battaglia2018relational, schulz2012finding}. This structured form embodies an inductive bias at a specific level of abstraction and may seem appropriate to capture human knowledge in one particular domain and task over the other. For instance, a graph structure most intuitively represents causal relationships, but probabilistic graph grammars are at a higher level of abstraction and can specify rules for generating these graph structures. To represent concepts universally, we need a structural inductive bias that not only supports rapid learning and generalization of concepts from a few examples, but also captures the nature of concepts itself.

Concepts have some inherent structure and reflect the world. The form includes parts, sub-parts, and relations, and these modules scale up to general concepts when combined in different ways. General concepts can themselves be organized in various structures such as hierarchical, network, or ring-like forms to create meaning. The ability to compose simple concepts into complex ones in a purposeful way supports the boundless nature of thought \cite{fodor1975language, goodman2014concepts}. Concepts can also be assimilated into a single module referred to by a name instead of its comprising components. For instance, anytime one needs to apply the concept of adding two natural numbers, they can directly access a representation of "addition" which combines individual concepts of "natural numbers," "counting" of two instances of numbers, and the "plus" operation over those two instances. It is also possible that the meaning of a concept can be updated over time. For example, the "addition" concept only concerns natural numbers initially but can be updated later to support all "real" numbers. Moreover, concepts attribute meaning by virtue of the causal relations within the concept or between other concepts \cite{piantadosi2021computational}. For example, "plus" is a common arithmetic symbol that becomes meaningful because of how it is used in relation to the numbers it is applied to and the result it produces: plus causes the input numbers to be added together to result in a total of those numbers. Lastly, there is an implicit uncertainty in the meaning a concept can take on. One concept can have several interpretations. For example, the concept of a "horse" encapsulates all instances of horses invariant of their breed, size, and color. Thus, a universal concept representation should support modularity, abstraction, temporal state change and embody causal and probabilistic traits. 

Programs are a structured representation which provide the necessary scaffolding for universally representing concepts. In the following sections, we define what a program is and show how it satisfies the representational demands presented above. 

\section{What structural form does a program representation take?}

A program is a sequence of symbols that are syntactically ordered according to a formal language to semantically represent instructions that a computational system can execute. For instance, the following piece of code represents program written for addition in the Python programming language:
\begin{verbatim}
    1 def addition(x, y):
    2    return x + y
\end{verbatim}

It contains symbols "\texttt{+ ( , ), :}" and terms like "\texttt{def, addition, return, x, y}." Semantically, "\texttt{def}" suggests a function definition, "\texttt{addition}" describes the name and purpose of the function, the "\texttt{( , )}" symbols encapsulate the input parameters that one would generally pass to the function, and the "\texttt{:}" symbol marks the beginning of the function body. The "return" symbol indicates that the following statement must be the output, and symbols "\texttt{x, +, y}," in that order, define the output as the addition of values contained in the variables \texttt{x} and \texttt{y}. Such semantic and syntactic rules are generally specified by a formal language in terms of regular or context-free grammars. 

A grammar gives syntactical meaning to a sequence of symbols of a program in terms of a set of tokens, non-terminals, a start symbol and a set of production rules. Below is an example of a the grammar underlying the python program for a function definition \cite{CS-R9526}.

\begin{verbatim}
    1 function_def_raw:
    2    | 'def' NAME '(' [params] ')' ['->' expression ] 
    3        ':' [func_type_comment] block 
    4    | ASYNC 'def' NAME '(' [params] ')' ['->' expression ] 
    5        ':' [func_type_comment] block 
\end{verbatim}

Lines 1-5 describe the rule for generating expressions of function definitions. This rule can be generally applied to express any function definition in Python. 



"Context-free" grammars (CFG) are the most widely accepted formalism which allows production rules to be applied regardless of which symbols surround a non-terminal. In contrast, "context-sensitive" grammars are more general and impose dependency between the context surrounding terminals and non-terminals between the left and right-hand sides of the production rules. Probabilistic Context-Free Grammars extend CFGs by defining a distribution over these possible derivations. Often, researchers assign a prior probability to the PCFG such that shorter and simpler productions are more likely \cite{goodman2008rational}. This design choice is congruent with a common principle in Computer science called \textit{Occam's razor}, which posits that the simplest hypothesis explaining some given data should always be preferred. A closely related principle, \textit{minimum description length}, formally defines this notion in terms of \textit{Kolmogorov complexity}, which is the length of the program that explains the data. It says that one should choose the shortest program that explains the data. Thus, suitable priors over PCFGs are those that select shorter expressions. 

One can also write a Python program to express other structured representations, such as graphs, as shown below. This suggests that programs are a super-set of other structured representations. 

\begin{verbatim}
    1 class Graph:
    2    def __init__(self, nodes, edges):
    3        self.adjacencyList = [[] for _ in range(nodes)]
    4        for (from, to) in edges:
    5            self.adjacencyList[from].append(to)
\end{verbatim}

 The same can be achieved in another programming language, but the meaning of the grammar will remain fairly consistent. For example, below is the grammar for function definitions in C, semantically analogous to the Python grammar introduced earlier, but with slight differences capturing syntactic nuances \cite{ritchie1988c}. 

\begin{verbatim}
    1 func: type id '(' parm_types ')' 
    2           '{' { type var_decl { ',' var_decl } ';' } 
            { stmt } '}'
    3     | void id '(' parm_types ')' 
    4           '{' { type var_decl { ',' var_decl } ';' } 
                { stmt } '}
\end{verbatim}
 
 Program representations have several additional  properties that are useful for representing concepts and operating over them. In the following section we will make a case for programs as an ideal choice for universally representing concepts.

\section{What makes programs a good candidate for a universal representation?}

In the previous sections, we outlined the required properties of a concept representation and suggested programs as a viable candidate. In this section, we specify how programs satisfy those properties.


\paragraph{Modularity} is possible because modules of programs can themselves can be separated or pieced together to create different meanings. For example, a function for computing "\texttt{mx + b}" can make calls to helper functions "\texttt{multiplication}" and "\texttt{addition}", or a group of functions representing individual concepts can be grouped together into a class, i.e., "\texttt{class Dog}" can include member functions to represent the concepts of "\texttt{breed}" and "\texttt{age}" that attribute a dog its character. This way of grouping, slicing, and arranging concepts in specific ways is pervasive within the conceptual universe.

\paragraph{Abstraction} is implicit in programs because individual expressions which have specific meanings can be grouped as a single module, e.g., a variable, function, class, file, or library, to attribute higher-level meaning. The availability of abstraction in programs is valuable because concepts are described in terms of other concepts such that internal concepts are not re-specified but instead accessed from memory at a higher level of abstraction. 

\paragraph{State change} of a primitive variable in a program occurs when the value of that variable gets overwritten during program execution. Programs are great at capturing similar dependencies in learning concepts: a concept may initially have some meaning but can become more specific or broad as new concepts are learned. 

\paragraph{Causality} is embedded in programs because one must not only require that the program expression generates the correct outcome but also consider the correct causal influence of all the possible interventions one can make to that expression \cite{chater2013programs}. For example, a function definition encodes how the input parameters causally influence the function's output. Thus, programs help model the causal structures within a concept or between concepts. 

\paragraph{Uncertainty} is natural in the world of programs. There can be multiple valid expressions to represent the same concept. For example, to create a list of objects, one can initialize an empty list data structure and define a for-loop that appends objects to the list, or alternately, one can use list comprehension to do the same in one line. On the other hand, probabilistic programs allow the same program to generate multiple instances of a concept. Uncertainty is valuable for representing and learning concepts because one concept can generate multiple examples of itself, but also, a single concept can have varying interpretations based on context.

\section{Is representation enough for universality?}

Although programs are a strong candidate for a universal representation of concepts, it is unclear whether one can disentangle the representation from its algorithm for efficiently learning concepts under limited cognitive resources. In principle, if our knowledge of concepts is indeed encoded as a program, we can run the program forward to generate new instances of a concept. Then learning the concept is inferring, or synthesizing a program for that concept \cite{rule2020child}. 

Previously, we established that human infants access structured mental representations to impose an inductive bias over their search space. Suppose the human mind uses program representations as an inductive bias. How does the mind choose suitable priors over programs, or in other words, which concepts are defined primitively in the program representation and which concepts are learned? These questions are important as they influence the initial setup of the program representation in terms of what primitives to include in the grammar. 

Computability theory has shown that there are a class of concepts that can be learned in polynomial time using Conjunctive Normal Form (CNF) expressions \cite{valiant1984theory}. However, it does not elaborate on how concepts outside of this class can be learned the way humans broadly do. Researchers have also investigated how humans induce programs \cite{sanborn2018representational}. However, it is unclear whether humans begin with intuitive knowledge or start with a clean slate \cite{lake2017building}. 

Despite the algorithmic challenges highlighted above, several works, including computational modeling work in concept learning, have proposed how to search within a space of programs efficiently \cite{ellis2018library, ellis2020dreamcoder, valkov2018houdini}. In the following section, we evaluate the different formats in which programs have been represented in the field and analyze their contribution toward a universal representation of concepts. 


\section{Various ways concepts have been represented as a program in computational models of concept learning}

A large body of work in cognitive science assumes that programs are the best general purpose form of structured representations of human knowledge. There are two main ways people have used programs to represent concepts. The first way treats concepts as a set of its instances where each instance is a program defined by a grammar. For example, the concept of blueness is the set of all programs in some grammar that corresponds to blue objects. The second way treats concepts as a set of interdependent programs. For example, the concept of colors blue, green and yellow each have a corresponding program definition, where the program for green is consistent with the concept that mixing the colors blue and yellow create green. In this section, we survey the various formats in which programs were used by grouping them by these two categories. Then we assess their contribution towards a universal representation of concepts by weighing the strengths and weaknesses of each approach. 







\subsection{Concept as a set of all its program instances}
One class of representation views concepts as a set of all its instances. This set is defined by rules and each instance of the set is a program expression generated by those rules. The rule discriminates between categories of objects based on features or relations \cite{kemp2012exploring}. For example, one understands the concept of a horse by discriminating a set of horse instances from a set of non-horse instances based on certain attributes such as ears, eyes, tail, and mouth. The production rules of the grammar provide a useful inductive bias for searching an infinite space of hypotheses as it creates clear discriminatory boundaries within that space based on certain properties of those instances. The works surveyed in this section all represent concepts as grammars, but differ in terms of the semantic and syntactic design choices of the rules. This influences the hypothesis space of a concept learner because only semantically consistent concepts will be available in the hypothesis space and only syntactically valid instances of the concepts can be generated. 
 
In the simplest case, rules have been defined as \textbf{boolean functions} that map objects to truth values, where an object is true if it belongs to a concept. Following this view, \citet{goodman2008rational} represent concepts as a structured hypothesis space of logical rules generated by a probabilistic grammar called the "concept language of logical rules". Their representation assumes that each concept corresponds to a single rule and restricts the space of concepts to logical connectives such as $\wedge$, $\vee$, $\Rightarrow$, quantifiers such as $\forall x$, and function and comparison feature predicates $f_i(x)$, $=$, $<$ and $>$, over an object $x$. Bayesian learning is a common algorithm adopted in concept learning to generalize and learn from a small set of examples \cite{tenenbaum2001generalization}. Based on this, they propose a Rational Rules model which performs Bayesian inference over the representation to predict the rule that explains a concept. 

In the case of Bongard problems \cite{bongard1970pattern}, a visual task over a class of \textbf{geometric shapes}, a concept is also viewed as a set of all its instances. A typical Bongard problem presents two concepts side by side in terms of six of its instances. There is one simple rule that discriminates the instances from the left class and the right class. \citet{depeweg2018solving} represent Bongard problems as Probabilistic Context Free Grammars so that a single rule can correspond to a concept and instances of the concept can be probabilistically sampled by this rule. This work also adopts Bayesian inference for learning visual geometric concepts. 

\begin{figure}
    \centering
    \includegraphics{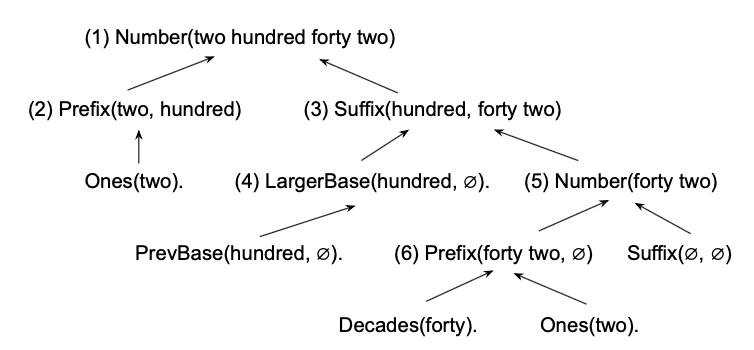}
    \caption{Example of Range Concatenation Grammar proposed by \citet{rule2015representing} for the Number phrase \textit{two-hundred-forty-two}. The grammar is context-sensitive and supports the compositional and relational nature of number knowledge concepts.}
    \label{fig:prcg}
\end{figure}

So far, we have treated concepts as sets that are clearly separated by its instances. However, there are certain concepts like natural numbers which cannot be learned in isolation this way because the set of instances is infinite \cite{rule2015representing}. Moreover, some concepts are learned in terms of other concepts, e.g. "left" is understood in terms of other directions like "up", "down", and "right. \citet{rule2015representing} address this by representing concepts as networks of possible \textbf{relations} in the domain of number knowledge. These networks are expressed as a Probabilistic Range Concatenation Grammar (PRCG) which is a context-sensitive grammar with a prefix-base-suffix system that specifically satisfies the relational requirement Figure~\ref{fig:prcg}. Their grammar restricts the hypothesis space by defining primitive predicates: \texttt{number, succ, pred}, and relations: \texttt{more, less}. Their concept learning model performs Hierarchical Bayesian inference to learn the distribution over the parameters of a Latent Predicate Network structure which is defined in terms of a hierarchy of layers of \textit{observed}, \textit{latent} and \textit{lexicon} predicates, in that order. 

\citet{kemp2012exploring} proposes a general compositional representation language in terms of a grammar over \textbf{predicate logic} for a wider range of concept domains. The language contains primitives such as "for all" and "there exists" to make statements like "all squares have a slash". The conceptual universe is structured in terms of \texttt{domain, concept type, item, objects, features} and \texttt{relations}. Each domain corresponds to a set of concept types. Each concept type is a set of items grouped according to a systematic rule over objects, features and relations. In this work too, a concept acts as a discriminator function over a set of items based on some high-level feature. Here features can be additive which describe whether a feature is present or absent, or substitutive which describes whether size is big or small. Their model is defined as a search problem such that the learner has to search for a minimum description (or shortest rule) of the concept over a space of possible descriptions. 

An alternate view presents rules as generative models of concepts called \textbf{probabilistic programs} \cite{lake2015human, overlan2017learning, hwang2011inducing}. Previously the universe of concepts was structured in terms of objects, features and relations \cite{kemp2012exploring}. Here, the conceptual universe is structured compositionally in terms of parts, sub-parts and spatial relations. Each of these levels is a generative model represented by a probabilistic program such that we can sample new types of concepts and a specific type of concept can itself generate new examples (tokens) of itself. \citet{lake2015human} focus on a task set of visual concepts called Omniglot containing 1623 handwritten characters (images and penstrokes) from 50 writing systems. They propose a Bayesian Program Learning method for one-shot learning of Omniglot concepts and show that the model achieves human-level accuracy and outperforms common deep learning models \cite{krizhevsky2012adv, lecun1989backpropagation, salakhutdinov2012learning}. 

\begin{figure}
    \centering
    \includegraphics{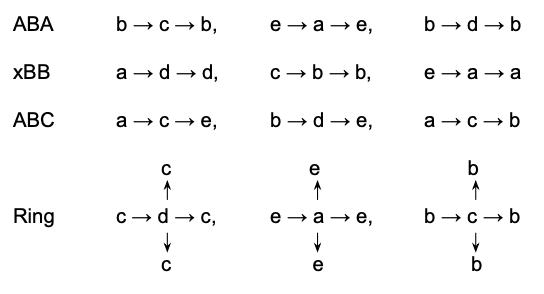}
    \caption{Example of visual concepts generated by \citet{overlan2017learning}. Concepts are represented as rules \textit{ABA, xBB, ABC, Ring} which sample objects \textit{a, b, c, d, e} in order.}
    \label{fig:rule_example_overlan2017learning}
\end{figure}

In previous sections, we acknowledged that abstraction is a key requirement for representing concepts. Typically, this abstraction is achieved via variable binding which allows a human concept learner to assign a name to a piece of information for easily storing and retrieving from memory. Using variable names allow concepts to be abstract because we can refer to concepts in terms of their abstract names instead of the specific details that make up the concept. Rules can be defined in terms of these variable names to demonstrate the relationships between abstract concepts instead of the features of concepts themselves. \citet{overlan2017learning} express rules for generating abstract concepts like \textit{ABA, xBB, ABC, Ring} as a Probabilistic Context-Free Grammar (PCFG) in terms of \textbf{lambda calculus} expressions with variable bindings. Each concept is an object made up of three-individual parts \textit{a, b, c, d, e} and a generative rule which arranges parts in an order by randomly sampling a part without replacement Figure~\ref{fig:rule_example_overlan2017learning}. A human learner shows causal knowledge of the origin of the concepts by inferring the correct rule, and use it to classify new objects into the appropriate concepts. They propose a "Hierarchical Language of Thought" model which computes a posterior distribution over hypotheses using a Markov Chain Monte Carlo inference procedure. 

In one work, \citet{mao2019neuro} present concepts in the context of a visual question and answering task. In this case, concepts contain objects, features and relations and are still viewed as a set of all its instances, but are expressed as a deep latent representation from a perception module. The questions and answers, however, are represented as a \textbf{domain-specific language (DSL)} which customizes the grammar for their visual domain. The DSL contains primitives such as "\texttt{Scene, Filter, Relate, Union, Intersection, Exist, Count}" as functional rules. Another work represents concepts as "cognitive programs" in a tabletop world in which objects are spatially arranged on a tabletop and concepts capture the relations between these objects \cite{lazaro2019beyond}. Here, concepts are tightly coupled with a vision, attention and control system for a domain of visual concepts. The program representation is also a domain specific language capturing a sequence of imperative instructions such as "\texttt{scene\_parse(), top\_down\_attend(), grab\_object(), move\_hand\_to\_fixation(), change\_color()}". The authors present a program induction model over this representation and show that robots embodied with this system can understand concepts from schematic drawings and apply them to tasks, just like humans. However, unlike previous Bayesian models which use small datasets, this model was trained on 546 input-output examples of concepts and their respective programs. Even \citet{mao2019neuro} propose a successful model but include 5000 images in their training dataset. Nevertheless, these works are interesting because they present an alternate way to restrict a learner's hypothesis space by carefully choosing the primitives, syntax and semantics via a DSL. Due to this, learning in the hypothesis space becomes more efficient, but the language becomes less generic and rigid. In contrast, humans are able to flexibly learn concepts over a broad range of domains and apply them universally with very few examples. 


All of these works represent concepts as a rules that can generate a set of instances, but differ semantically based on the class of concepts they seek to capture. These semantic differences are plausible because they address the algorithmic concern of exploring an infinite search space; focusing on a specific class of concepts allows us to propose more tractable models of concept learning. Ultimately, a universal representation of concepts should be able to handle all possible classes of concepts. In the following section, we review an alternate representation that treats concepts as a set of interacting programs. 

\subsection{Concept as a set of interdependent programs}

A second class of representation views concepts as a set interdependent programs \cite{piantadosi2021computational, liang2010learning, dechter2013bootstrap}. The representation is defined in terms of \textbf{Combinatory Logic (CL)}, a variable-free subset of lambda calculus, which contains primitive combinator functions \texttt{S, K, I}. No domain or class specific primitives are required; in principle, all concepts can be constructed out of an initial set of combinators. A concept gets its meaning by virtue of how it is used in a context. For example, the concept \textit{blue} can mean the color "blue" or could denote "melancholy" if used in a context referring to human emotions. Concepts can be easily expressed in terms other concepts because sub-expressions are themselves well-formed and can be shared or reused. For example, the concept of the color \textit{purple} can be defined in terms of the concepts \textit{blue} and \textit{red}. If one lacks domain knowledge for a certain class of concepts, CL can be especially useful because it can build up the knowledge by reusing and composing learned concepts. These qualities make combinatory logic a competitive candidate for a generic universal representation as it assumes the least inductive bias over the search space. \texttt{S\&K} are two common combinators in CL. The \texttt{S} combinator is defined as a function that takes three functions $x$, $y$, and $z$, as inputs, passes $z$ to $x$ and $y$ each and composing the results:  \texttt{$(\textbf{S} \ x \ y \ z) \rightarrow ((x \ z) (y \ z))$}. The combinator K is defined as a function that takes in two inputs $x$ and $y$ and ignores $y$: \texttt{$(\textbf{K} \ x \ y) \rightarrow (x)$}.  \citet{piantadosi2021computational} makes a case for CL as a universal mental representation by showing how a wide range of concepts can be represented by composing just two combinators, \texttt{S\&K}. For example, if \texttt{K} represents the concept \textit{false} and \texttt{(K K)} represents the concept \textit{true} then \texttt{((S \ S)(K (K \ K)))} is a possible representation for the concept \textit{or}. Applying the combinator for \textit{or} to \textit{true} and \textit{false}, \texttt{(((S \ S) (K (K \ K))) (K \ K) K)} correctly produces \texttt{(K K)} which represents \textit{true}. 

Although \texttt{S$\&$K} is sufficient to describe any function universally, the search space can become quite large and difficult to interpret by a human reader. \citet{liang2010learning} modify \texttt{S$\&$K} by adding primitive combinators called routers \texttt{B, C, S} which route the computation to the right sub-tree, left sub-tree or both left and right sub-trees respectively. The combinator expressions are represented by a Probabilistic Context-Free Grammar 
The authors of this work highlight that programs that share common subprograms sometimes need to be refactored because they do not always explicitly reflect semantic similarities. They allow some combinators to be refactored by others by defining transformations based on the Metropolis-Hastings algorithm. Their representation is tested in arithmetic and text editing domains. The arithmetic domain contains primitive combinators for arithmetic operations + and -, and comparison operations > and <. The text editing domain contains primitive combinators for  \texttt{string-append, cursor position delete, copy, cut, move}, and \texttt{find}. A follow-up work \cite{dechter2013bootstrap} adopts the same representation but include additional primitive combinators to satisfy domain-specific requirements for two domains, Boolean function learning, which requires the logical primitive NAND, and symbolic regression, which requires arithmetic primitives \texttt{1, 0, *, +}. They represent the combinators in the form of stochastic grammars by defining a distribution over binary trees and applying a polymorphic type system. 

One limitation of \citet{piantadosi2021computational}'s work is that \texttt{S\&K} representations are recomputed for each domain and cannot be reused. \citet{liang2010learning} make up for this by showing that their representation can be applied to multi-task learning. However, the multiple tasks considered in this work share a common set of primitives but it is possible that some concepts have their own private primitives. 
Recent work has shown how concepts can be learned across domains that use different set of primitives \cite{ellis2018library, ellis2020dreamcoder}. 



In principle, any concept can be built up using CL expressions. However, in practice, since the search space is so large, researchers ultimately restrict the search space by testing the representation in a specific domain. Thus, it is unclear whether CL representation can be disentangled from its algorithmic use. 


\subsection{What is the overarching gap?}

Representing concepts as a set of instances is useful because concepts can be treated as generative models. In the forward direction, instances of a concept can be sampled, and in the reverse direction, the rule defining the concept can be inferred. However, this representation requires a set of primitive terms to be defined. In contrast, representing concepts as a set of interdependent programs does not require primitives as it can be built up from a basic set of combinators or other concepts. Nevertheless, this representation is deterministic and fails to capture the notion of instances of a concept. 

A universal concept representation should ideally combine the powers of both perspectives. One interesting direction for future work can explore how concepts can be represented using combinatory logic as a probabilistic program for learning concepts over multiple domains. Since it is unclear whether representation can be disentangled from its algorithmic use, another direction of research can test the value and extent of representation in algorithmic efficiency. 

\section{Conclusion}
Programs are unarguably a good candidate for a universal representation of concepts. Despite the representational and algorithmic challenges presented above, a large body of work has demonstrated their representational power. In general, programs have been used to express concepts in two distinct ways: (1) as an instance among a set of all instances of a concept and (2) as interacting entities of a concept. Ultimately, both representations can be combined to leverage their individual power. 

\begin{ack}
I thank Todd Millstein for his close guidance and advice throughout this work. I also thank Peter Battaglia for his mentorship during my M.S. program and for the guidance that helped shape this work.
\end{ack}

\bibliography{references}
\end{document}